\documentclass[11pt]{article}

\usepackage[preprint]{acl}

\usepackage{times}
\usepackage{latexsym}

\usepackage[T1]{fontenc}

\usepackage[utf8]{inputenc}

\usepackage{microtype}

\usepackage{inconsolata}

\usepackage{graphicx}

\usepackage{amsmath}
\usepackage{booktabs}
\usepackage{multirow}
\usepackage{multicol}
\usepackage{amssymb}
\usepackage{algorithm}
\usepackage{algorithmic}
\usepackage{xcolor}

\newcommand{\TFred}[2]{#1 {\scriptsize \textcolor{red}{(#2\%)}}}

\title{Reliability-Aware Adaptive Self-Consistency for Efficient Sampling in LLM Reasoning}

\renewcommand{\thefootnote}{\fnsymbol{footnote}}

\author{%
Junseok Kim$^1$ \quad Nakyeong Yang$^1$ \quad Kyungmin Min$^1$ \quad \textbf{Kyomin Jung}$^1$\footnotemark[2]\\
$^1$IPAI, Seoul National University\\
\texttt{\{kim.junseok,yny0506,kyungmin97,kjung\}@snu.ac.kr}
}

\begin{document}
\maketitle
\footnotetext[2]{Corresponding author.} 
\renewcommand{\thefootnote}{\arabic{footnote}}
\begin{abstract}
Self-Consistency improves reasoning reliability through multi-sample aggregation, but incurs substantial inference cost.
Adaptive self-consistency methods mitigate this issue by adjusting the sampling budget; however, they rely on count-based stopping rules that treat all responses equally, often leading to unnecessary sampling.
We propose \textbf{Re}liability-Aware \textbf{A}daptive \textbf{S}elf-\textbf{C}onsistency (\texttt{ReASC}), which addresses this limitation by reframing adaptive sampling from response counting to evidence sufficiency, leveraging response-level confidence for principled information aggregation.
\texttt{ReASC} operates in two stages: a single-sample decision stage that resolves instances confidently answerable from a single response, and a reliability-aware accumulation stage that aggregates responses by jointly leveraging their frequency and confidence.
Across five models and four datasets, \texttt{ReASC} consistently achieves the best accuracy-cost trade-off compared to existing baselines, yielding improved inference efficiency across model scales from 3B to 27B parameters.
As a concrete example, \texttt{ReASC} reduces inference cost by up to 70\% relative to self-consistency while preserving accuracy on GSM8K using Gemma-3-4B-it.
\end{abstract}

\section{Introduction}
Large language models (LLMs) have demonstrated strong performance on complex reasoning tasks, yet the inherent stochasticity of decoding introduces variability in intermediate reasoning trajectories, making it difficult to reliably obtain a correct reasoning trajectory from a single generation.
Self-Consistency (SC) addresses this challenge by sampling multiple reasoning paths and aggregating their final answers, effectively accumulating evidence and yielding consistent performance gains \cite{wang2022self}.
\begin{figure}[t]
\centering
\includegraphics[width=\linewidth]{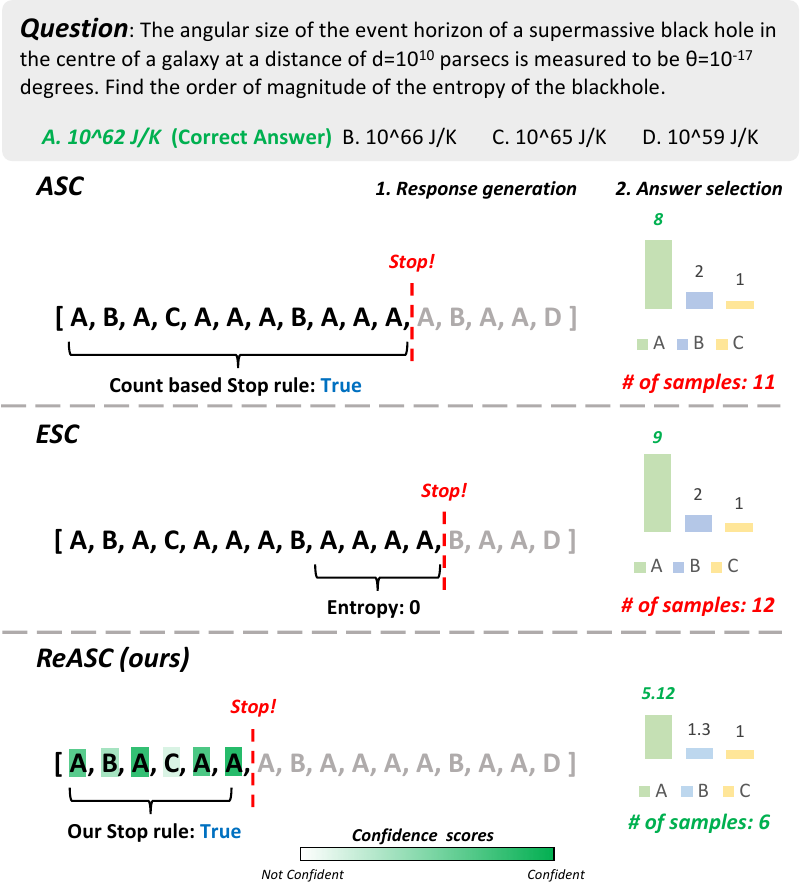}
\caption{\textbf{Count-based stopping may lead to inefficient evidence accumulation.}
Ignoring response reliability, count-based criteria may require unnecessary additional samples, while \texttt{ReASC} reaches the same decision with fewer samples.
}
\label{fig:failure_case}
\vspace{-0.45cm}
\end{figure}
However, SC relies on a fixed sampling budget, applying the same number of samples to all inputs.
As a result, some instances continue sampling even after sufficient evidence has already been accumulated, while others remain unresolved even after the sampling budget is exhausted.
To mitigate this inefficiency, adaptive self-consistency variants such as Adaptive Consistency (ASC) \cite{aggarwal2023let} and Early-Stopping Self-Consistency (ESC) \cite{li2024escape} dynamically adjust the sampling budget based on observed responses.
These methods primarily rely on count-based criteria to guide sampling decisions.

From an evidence accumulation perspective, this reliance on count-based aggregation treats all sampled responses as equally informative.
By design, such aggregation does not account for differences in response reliability.
However, reasoning trajectories generated under stochastic decoding can vary in reliability, with some responses providing strong evidence while others being noisy or misleading.
As a result, early reliable signals can be diluted by later unreliable responses, making it difficult to recognize when sufficient evidence has already been accumulated.
This failure mode is illustrated in Figure~\ref{fig:failure_case}, where ASC and ESC continue sampling even though the accumulated responses already provide sufficient evidence for a reliable decision.

These observations suggest that adaptive sampling decisions should be guided not only by how often an answer appears, but also by how much reliable evidence each individual response contributes to the final decision.
Such response-level reliability can be captured from model confidence signals during generation, which provide instance-specific information about how strongly the model supports a given response \cite{wang2024chain, wang2024soft}.
Among various confidence signals, \textit{self-certainty} has been shown to correlate with the reliability of reasoning trajectories, making it a suitable basis for guiding evidence accumulation without additional supervision \cite{kang2025scalable}.

Building on this insight, we propose \textbf{Re}liability-Aware \textbf{A}daptive \textbf{S}elf-\textbf{C}onsistency (\texttt{ReASC}), an adaptive self-consistency framework that incorporates a \textit{self-certainty} variant as a response-level reliability signal to guide how evidence is accumulated at inference time.
\texttt{ReASC} decomposes inference into two complementary stages, separating instances by evaluating the evidence sufficiency for each instance.
In the first stage, the model evaluates the confidence of a single response to determine whether sufficient evidence is already available for a reliable decision.
If additional evidence is required, the second stage performs reliability-aware evidence accumulation, allowing high-confidence responses to contribute more evidence than low-confidence ones.
By guiding sampling decisions based on confidence-weighted evidence sufficiency rather than response counts alone, \texttt{ReASC} makes reliable decisions with fewer samples.

Empirically, \texttt{ReASC} consistently outperforms existing adaptive sampling baselines in the accuracy-cost trade-off across five models from three major families (LLaMA, Qwen, and Gemma) and four datasets.
For instance, on GSM8K with Gemma-3-4B-it, \texttt{ReASC} reduces inference cost by approximately 70\% relative to SC, while maintaining accuracy over prior adaptive baselines.
Further analysis reveals that this efficiency gain arises from the complementary roles of \texttt{ReASC}'s two stages.
The first stage correctly resolves a substantial fraction of instances accurately with a single response.
Notably, for instances that proceed beyond the first stage, the second stage still achieves substantial cost reductions without compromising accuracy.
Together, these results show that \texttt{ReASC} establishes a principled framework for adaptive sampling by jointly modeling response counts and response reliability.
Our contributions are as follows:
\begin{itemize}
    \item We propose \texttt{ReASC}, a reliability-aware adaptive self-consistency framework that accumulates evidence by jointly considering response counts and response-level reliability.
    \item We empirically characterize the limitation of count-based stopping criteria, showing that treating all responses as equally informative can lead to unnecessary additional sampling.
    \item Through experiments and analyses, we show that \texttt{ReASC} consistently reduces inference cost while maintaining accuracy through the complementary roles of each stage.
\end{itemize}

\section{Related Works}
\paragraph{Adaptive Sampling for Self-Consistency.}
Self-Consistency (SC) improves reasoning reliability by sampling multiple reasoning trajectories and aggregating via majority voting, but relies on a fixed sampling budget, resulting in substantial inference cost \cite{wang2022self}.
To address this limitation, adaptive self-consistency variants have adjusted the sampling budget based on observed responses \cite{aggarwal2023let, li2024escape, wang2025make}.
While these approaches reduce inference cost compared to SC, they rely on stopping criteria based on answer frequency or agreement patterns, implicitly treating all sampled responses as equally informative.
As a result, they often lead to inefficient sampling even when sufficient evidence already exists to make a sampling decision.

\paragraph{Confidence Estimation.}
A growing body of work studies confidence and uncertainty estimation in large language models to assess the reliability of generated predictions.
Prior work investigates response-level signals derived from model outputs, such as logit-based confidences and entropy-based uncertainty measures, showing that these signals correlate with prediction reliability \cite{kadavath2022language, kang2025scalable, zhang2023enhancing}.
Building on this line, several methods leverage confidence signals for post-hoc answer selection, reranking, and pruning of candidate responses \cite{wang2024soft, wang2024chain, taubenfeld2025confidence, fu2025deep}.
Our approach uses response-level confidence at inference time as a criterion for adaptive sampling.
Specifically, \texttt{ReASC} interprets confidence as an estimate of response reliability and uses it to modulate how evidence is accumulated across sampled responses.

\section{Preliminaries}
\paragraph{Confidence as Evidence Strength.}
\label{sec:self certainty}
Recent work has shown that confidence signals derived from a model's token-level probability distribution correlate with the reliability of its reasoning trajectories \citep{kang2025scalable, fu2025deep}.
In this work, we interpret response-level confidence as an indication of how reliable evidence a generated response provides.
Under this view, confidence naturally serves as a weighting signal in evidence accumulation, allowing more reliable responses to contribute more strongly.
We adopt \emph{self-certainty} as the underlying confidence signal \cite{kang2025scalable}.
Given the model's token probability distribution at decoding step $i$, the token-level self-certainty is defined as
\begin{equation}
    c_i = - \frac{1}{|\mathcal{V}|} \sum_{w \in \mathcal{V}} \log p(w \mid x, y_{\le i}),
\end{equation}
where $\mathcal{V}$ denotes the vocabulary.
Higher values correspond to a more concentrated probability distribution over tokens, reflecting greater model confidence during generation.
Following \citet{kang2025scalable}, response-level self-certainty is computed as the average of token-level self-certainties over the reasoning trajectory.

\begin{figure}[t]
\centering
\includegraphics[width=\linewidth]{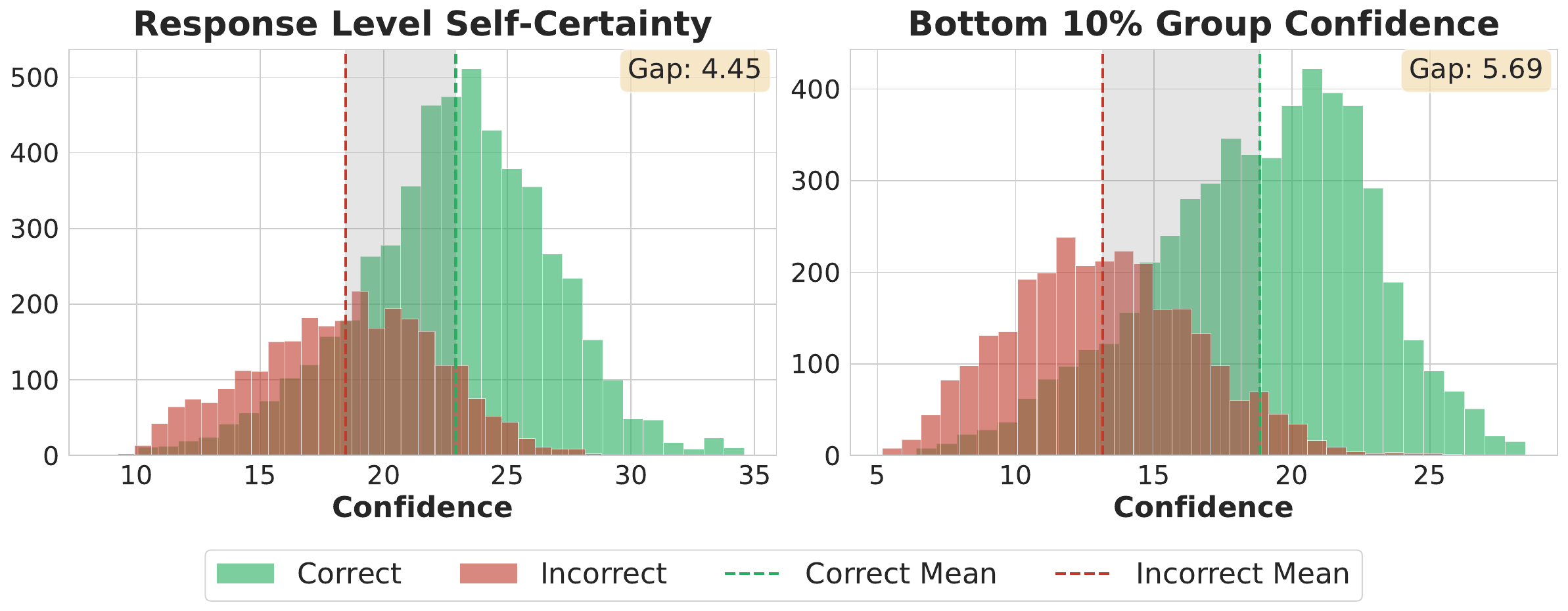}
\caption{
\textbf{Comparison of two confidence signals.}
Using Gemma~3~4B-Instruct on MATH500, Bottom~10\% Group Confidence shows a larger separation between correct and incorrect responses than Response-level Self-Certainty.}
\label{fig:confidence_comparison}
\vspace{-0.45cm}
\end{figure}

\paragraph{Bottom $10\%$ Group Confidence.}
\label{sec:bottom10}
Response-level self-certainty summarizes the average confidence of a generated response, but can obscure localized uncertainty within a reasoning trace.
To capture such localized uncertainty, we adopt the \emph{Bottom $10\%$ Group Confidence}.
Specifically, given a response $y$, the token sequence is partitioned into sliding-window groups $\{G_1, G_2, \ldots, G_n\}$ and the group confidence $C_{G_i}$ is computed by averaging the token-level self-certainties within the group $G_i$.
The Bottom $10\%$ Group Confidence is then defined as
\begin{equation}
\label{eq:bottom10}
C_{\text{bottom-}10}(y) = \frac{1}{|\mathcal{G}_{\mathrm{b}}|} \sum_{G_j \in \mathcal{G}_{\mathrm{b}}} C_{G_j},
\end{equation}
where $\mathcal{G}_{\mathrm{b}}$ denotes the set of groups with the lowest $10\%$ group confidence.
This metric emphasizes low-confidence segments indicative of unreliable reasoning, observed in prior work \citep{fu2025deep}.

To determine which confidence metric reliably reflects response quality, we compare Bottom $10\%$ Group Confidence with response-level self-certainty.
As shown in Figure~\ref{fig:confidence_comparison}, Bottom $10\%$ Group Confidence separates correct and incorrect responses more clearly than response-level self-certainty.
Accordingly, we use this metric as the confidence signal in \texttt{ReASC}, with additional analyses provided in Appendix~\ref{appendix:confidence_design}.

\section{Methods}
\begin{figure*}[t]
\centering
\includegraphics[width=\linewidth]{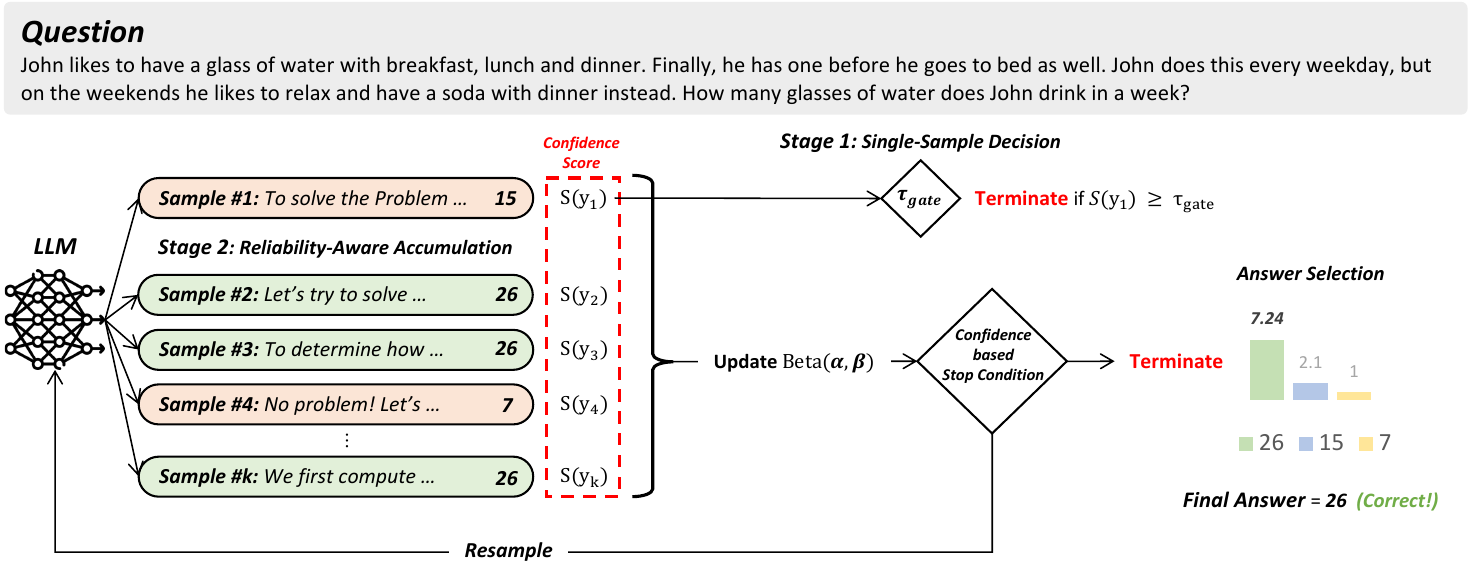}
\caption{
\textbf{Overview of \texttt{ReASC}.}
The model first attempts a Single-Sample Decision (Stage~1) by evaluating whether the response reliability is sufficient.
If not, it proceeds to Reliability-Aware Accumulation (Stage~2), where responses are adaptively sampled and aggregated via confidence-weighted Beta updates.
}
\label{fig:overview}
\vspace{-0.45cm}
\end{figure*}
\label{sec:overview}
\texttt{ReASC} is a reliability-aware adaptive self-consistency framework that decomposes inference into two stages to enhance efficiency.
In \textbf{Stage~1 (Single-Sample Decision)}, the model evaluates the confidence of a single response to assess whether a reliable decision can be made without additional evidence, thereby avoiding unnecessary evidence accumulation.
Inputs requiring additional evidence proceed to \textbf{Stage~2 (Reliability-Aware Accumulation)}, where the model accumulates confidence-weighted evidence from additional responses until a reliable decision can be made.
An overview of \texttt{ReASC} is shown in Figure~\ref{fig:overview}.

\subsection{Stage 1: Single-Sample Decision}
\label{sec:stage1}
In Stage~1, \texttt{ReASC} assesses whether a reliable decision can be made from a single response.
The model generates a response $y$ and computes a response-level confidence score $S(y)$ using the Bottom~$10\%$ Group Confidence defined in Equation~\ref{eq:bottom10}:
\begin{equation}
S(y) = C_{\text{bottom-}10}(y),
\end{equation}
which is interpreted as an estimate of response reliability and compared against a data-calibrated gating threshold $\tau_{\text{gate}}$.
If $S(y) \ge \tau_{\text{gate}}$, the response is accepted as providing sufficiently reliable evidence to determine the answer, and no further sampling is performed.
Otherwise, the instance proceeds to Stage~2 to accumulate additional evidence.
The selection of $\tau_{\text{gate}}$ is detailed in Section~\ref{sec:threshold}.

\subsection{Stage~2:~Reliability-Aware~Accumulation}
\label{sec:stage2}
When additional evidence is required, \texttt{ReASC} enters Stage~2 and accumulates confidence-weighted evidence from additional responses, interpreting confidence as an estimate of response reliability.
Rather than relying on count-based agreement, this stage evaluates whether the accumulated evidence is sufficient to make a reliable decision.

\paragraph{ASC Beta Update.}
We first review the Beta-based stopping rule used in Adaptive Consistency (ASC) \citep{aggarwal2023let}.
Let $V$ denote the set of sampled responses, and let $v_1$ and $v_2$ be the counts of the most frequent and second most frequent candidates in $V$.
The stopping decision is formulated as a binary comparison between these candidates, yielding a Beta posterior
\begin{equation}
p \sim \mathrm{Beta}(\alpha,\beta), \qquad \alpha = v_1 + 1,\; \beta = v_2 + 1,    
\end{equation}
where $p$ represents the probability that the most frequent candidate remains dominant as sampling continues.
ASC stops sampling when the $p$ exceeds a predefined threshold.

\paragraph{Confidence-Weighted Beta Update.}
The ASC formulation treats all sampled responses as equally informative, regardless of their reliability.
Incorporating reliability into the aggregation allows more informative responses to exert greater influence, enabling sufficient evidence to be recognized earlier without being dominated by less informative ones.
Building on this idea, we introduce a confidence-weighted variant of the Beta update, in which each sampled response contributes evidence that jointly accounts for its frequency and response reliability.
Given a confidence score $S(y_i)$, we standardize it using statistics $(\mu,\sigma)$ estimated from a held-out calibration set, denote the standardized confidence as $z(y_i)$, and map it to an evidence weight via an exponential transformation:
\begin{equation}
v(y_i) \leftarrow v(y_i) + \max(1,~\exp(\lambda z(y_i)))    
\end{equation}
where $\lambda$ controls the sensitivity of the confidence-to-evidence mapping.
This update can be interpreted as accumulating weighted pseudo-counts in the Beta posterior, allowing responses to contribute soft evidence proportional to their estimated reliability.
The exponential form amplifies high-confidence responses, while the $\max(1,\cdot)$ operation ensures a minimum unit contribution, maintaining compatibility with the original count-based formulation.
Among several reasonable designs, we find that this mapping yields stable stopping behavior across models and datasets (see Appendix~\ref{appendix:confidence_mapping}).
This confidence-weighted formulation preserves the ASC Beta posterior structure, where $(v_1, v_2)$ represent the accumulated confidence-weighted evidence of the two leading candidates.

\paragraph{Stopping Condition and Final Selection.}
Given the confidence-weighted Beta posterior, the stopping rule assesses the probability that the most frequent candidate remains dominant under further sampling.
Accordingly, for a $\mathrm{Beta}(\alpha,\beta)$ posterior, this probability admits the closed-form expression
\begin{equation}
\label{eq:probability}
P(p_1 > p_2 \mid V) = 1 - I_{1/2}(\alpha,\beta),    
\end{equation}
where $I_{1/2}(\alpha,\beta)$ denotes the regularized incomplete Beta function.
A detailed derivation of this expression is provided in Appendix~\ref{appendix:beta_derivation}.
Evidence accumulation continues until $P(p_1 > p_2 \mid V) \geq C_{\mathrm{threshold}}$, where $C_{\mathrm{threshold}}$ is a predefined confidence threshold, or until a maximum sampling budget is reached.
Once the stopping condition is met, the final answer is selected as $\hat{y} = \arg\max_y v(y)$, corresponding to the leading candidate supported by the accumulated confidence-weighted evidence.

\begin{algorithm}[t]
\small
\caption{Offline Gating Threshold Calibration}
\label{alg:offline-gating}
\begin{algorithmic}[1]

\REQUIRE Calibration set $\{(x_i, y_i^\ast)\}_{i=1}^k$, target accuracy $p_{\mathrm{target}}$
\ENSURE Gating threshold $\tau_{\mathrm{gate}}$

\STATE \textcolor{gray}{// (1) Compute confidence for each instance}
\FOR{$i = 1$ to $k$}
    \STATE Generate response $y_i$, then compute confidence $S(y_i)$
\ENDFOR

\STATE \textcolor{gray}{// (2) Compute mean confidence of correct responses}
\STATE $\mu_{\mathrm{correct}} \gets \mathbb{E}_{i:\,y_i = y_i^\ast}[\, S(y_i) \,]$

\STATE \textcolor{gray}{// (3) Compute accuracy-controlled threshold}
\STATE Sort $\{S(\hat{y}_i)\}$ in ascending order as candidate thresholds
\FOR{each threshold $t$}
    \STATE $V(t) \gets \{ y_i : S(y_i) \ge t \}$
    \STATE Compute $\mathrm{Acc}(t)$ over $V(t)$
    \IF{$\mathrm{Acc}(t) \ge p_{\mathrm{target}}$}
        \STATE $\tau_{\mathrm{accuracy}} \gets t$
        \STATE break
    \ENDIF
\ENDFOR

\STATE $\tau_{\mathrm{gate}} \gets \max(\mu_{\mathrm{correct}}, \tau_{\mathrm{accuracy}})$

\end{algorithmic}
\end{algorithm}
\subsection{Selecting Thresholds}
\label{sec:threshold}
\texttt{ReASC} leverages calibrated confidence statistics to support reliability-aware decision-making in both stages.
Specifically, the framework requires
(1) confidence statistics $(\mu,\sigma)$ to standardize response-level confidence scores for confidence-weighted Beta update in Stage~2, and
(2) a decision threshold $\tau_{\mathrm{gate}}$ that determines whether a reliable decision can be made from a single response in Stage~1.
Depending on the availability of labeled validation data, we estimate these quantities using one of two calibration settings: \textit{offline} or \textit{online} calibration.

\subsubsection{Offline Calibration}
\label{sec:offline_calibration}
In the offline setting, we assume access to labeled validation data and use a subset of $k$ labeled instances for calibration.
We compute the mean and standard deviation $(\mu,\sigma)$ of response-level confidence score from single-sample responses, which are used to standardize confidence in the confidence-weighted Beta update.

\paragraph{Selecting the Offline Gating Threshold.}
The gating threshold $\tau_{\mathrm{gate}}$ is designed to accept a single-sample response only when its reliability is sufficiently high.
Specifically, we consider two complementary criteria.
First, we compute the mean confidence of correctly solved instances, $\mu_{\mathrm{correct}}$, which reflects the typical confidence level of reliable single-sample decisions.
Second, to complement $\mu_{\mathrm{correct}}$, we derive an accuracy-controlled threshold $\tau_{\mathrm{accuracy}}$ that conservatively selects a high-confidence acceptance region by identifying the smallest confidence value $t$ such that the accuracy of instances accepted with $S(y) \ge t$ meets a target accuracy score $p_{\mathrm{target}}$.
The gating threshold is then defined as
\begin{equation}
\tau_{\mathrm{gate}} = \max\{\mu_{\mathrm{correct}},\ \tau_{\mathrm{accuracy}}\}.
\end{equation}
The full offline calibration procedure is summarized in Algorithm~\ref{alg:offline-gating}.
\begin{table*}[t]
\centering
\renewcommand{\arraystretch}{1.05}
\resizebox{\textwidth}{!}{
\begin{tabular}{
l l
c c c |
c c c |
c c c |
c c c
}
\toprule
& &
\multicolumn{3}{c}{\textbf{GSM8K}} &
\multicolumn{3}{c}{\textbf{MATH500}} &
\multicolumn{3}{c}{\textbf{Omni-Math}} &
\multicolumn{3}{c}{\textbf{GPQA-Diamond}} \\
\cmidrule(lr){3-5} \cmidrule(lr){6-8}
\cmidrule(lr){9-11} \cmidrule(lr){12-14}

\textbf{Model} & \textbf{Method} &
\textbf{Acc $\uparrow$} & \textbf{TF $\downarrow$} & \textbf{Acc/TF $\uparrow$} &
\textbf{Acc $\uparrow$} & \textbf{TF $\downarrow$} & \textbf{Acc/TF $\uparrow$} &
\textbf{Acc $\uparrow$} & \textbf{TF $\downarrow$} & \textbf{Acc/TF $\uparrow$} &
\textbf{Acc $\uparrow$} & \textbf{TF $\downarrow$} & \textbf{Acc/TF $\uparrow$} \\
\midrule

\multirow{6}{*}{\textsc{LLaMA-3.2-3B}}
& pass@1 & 73.09 & - & - & 41.4 & - & - & 11.7 & - & - & 17.26 & - & - \\
& SC (k=16)
& 83.93 & 12.31 & 6.82
& 54.8 & 25.31 & 2.17
& 16.1 & 44.99 & 0.36
& 25.38 & 30.62 & 0.83 \\
& ESC (w=4)
& 83.93 & \TFred{6.81}{-44.6} & 12.32
& 54.8 & \TFred{21.41}{-15.4} & 2.56
& 16.1 & \TFred{43.00}{-4.4} & 0.37
& 25.38 & \TFred{25.89}{-15.4} & 0.98 \\
& ASC
& 83.85 & \TFred{6.27}{-49.0} & 13.37
& 55.0 & \TFred{20.13}{-20.4} & 2.73
& 15.8 & \TFred{41.84}{-7.0} & \underline{0.38}
& 25.38 & \TFred{25.27}{-17.5} & \underline{1.00} \\
\cmidrule{2-14}
& \texttt{ReASC} (offline)
& 83.85 & \TFred{4.38}{-64.4} & \textbf{19.14}
& 55.0 & \TFred{18.27}{-27.8} & \underline{3.01}
& - & - & -
& - & - & - \\
& \texttt{ReASC} (online)
& 83.85 & \TFred{5.09}{-58.7} & \underline{16.47}
& 53.4 & \TFred{16.75}{-33.8} & \textbf{3.19}
& 15.3 & \TFred{34.90}{-22.4} & \textbf{0.44}
& 25.89 & \TFred{23.46}{-23.4} & \textbf{1.10} \\
\midrule

\multirow{6}{*}{\textsc{Qwen-2.5-3B}}
& pass@1 & 83.32 & - & - & 61.8 & - & - & 21.3 & - & - & 24.37 & - & - \\
& SC (k=16)
& 89.46 & 18.70 & 4.78
& 72.6 & 31.76 & 2.29
& 27.0 & 43.84 & 0.62
& 30.96 & 30.85 & 1.00 \\
& ESC (w=4)
& 89.39 & \TFred{8.03}{-57.0} & 11.13
& 72.6 & \TFred{21.75}{-31.5} & 3.34
& 27.0 & \TFred{38.58}{-12.0} & 0.70
& 30.96 & \TFred{22.46}{-27.2} & 1.38 \\
& ASC
& 89.39 & \TFred{7.57}{-59.5} & 11.80
& 72.6 & \TFred{20.84}{-34.4} & 3.48
& 27.2 & \TFred{36.69}{-16.3} & \underline{0.74}
& 30.96 & \TFred{21.26}{-31.1} & \underline{1.46} \\
\cmidrule{2-14}
& \texttt{ReASC} (offline)
& 89.46 & \TFred{5.70}{-69.5} & \textbf{15.69}
& 71.8 & \TFred{16.80}{-47.1} & \textbf{4.27}
& - & - & -
& - & - & - \\
& \texttt{ReASC} (online)
& 89.54 & \TFred{6.43}{-65.6} & \underline{13.93}
& 72.2 & \TFred{17.72}{-44.2} & \underline{4.07}
& 27.3 & \TFred{33.29}{-24.1} & \textbf{0.82}
& 30.96 & \TFred{19.88}{-35.6} & \textbf{1.56} \\
\midrule

\multirow{6}{*}{\textsc{Gemma-3-4B}}
& pass@1 & 88.40 & - & - & 63.2 & - & - & 25.2 & - & - & 21.32 & - & - \\
& SC (k=16)
& 92.12 & 32.67 & 2.82
& 71.6 & 50.15 & 1.43
& 29.9 & 90.36 & 0.33
& 30.46 & 52.74 & 0.58 \\
& ESC (w=4)
& 92.04 & \TFred{12.93}{-60.4} & 7.12
& 71.6 & \TFred{30.48}{-39.2} & 2.35
& 30.0 & \TFred{69.08}{-23.5} & 0.43
& 30.46 & \TFred{32.89}{-37.6} & 0.93 \\
& ASC
& 92.12 & \TFred{12.26}{-62.5} & 7.52
& 71.6 & \TFred{28.68}{-42.8} & 2.50
& 30.2 & \TFred{65.90}{-27.1} & \underline{0.46}
& 30.46 & \TFred{31.73}{-39.8} & \underline{0.96} \\
\cmidrule{2-14}
& \texttt{ReASC} (offline)
& 92.04 & \TFred{9.45}{-71.1} & \textbf{9.74}
& 71.4 & \TFred{25.59}{-49.0} & \textbf{2.79}
& - & - & -
& - & - & - \\
& \texttt{ReASC} (online)
& 92.04 & \TFred{10.25}{-68.6} & \underline{8.98}
& 71.4 & \TFred{26.17}{-47.8} & \underline{2.73}
& 30.5 & \TFred{62.20}{-31.2} & \textbf{0.49}
& 29.95 & \TFred{24.66}{-53.2} & \textbf{1.21} \\
\midrule

\multirow{6}{*}{\textsc{Qwen-2.5-7B}}
& pass@1 & 90.90 & - & - & 71.2 & - & - & 27.6 & - & - & 27.92 & - & - \\
& SC (k=16)
& 94.31 & 41.59 & 2.27
& 80.6 & 71.59 & 1.13
& 33.1 & 101.55 & 0.33
& 36.55 & 76.30 & 0.48 \\
& ESC (w=4)
& 94.24 & \TFred{13.87}{-66.7} & 6.80
& 80.6 & \TFred{39.71}{-44.5} & 2.03
& 33.1 & \TFred{83.97}{-17.3} & 0.39
& 36.55 & \TFred{49.54}{-35.1} & 0.74 \\
& ASC
& 94.24 & \TFred{13.40}{-67.8} & 7.04
& 80.8 & \TFred{37.25}{-48.0} & 2.17
& 33.3 & \TFred{80.24}{-21.0} & \underline{0.41}
& 36.55 & \TFred{45.63}{-40.2} & \underline{0.80} \\
\cmidrule{2-14}
& \texttt{ReASC} (offline)
& 94.09 & \TFred{10.43}{-74.9} & \textbf{9.02}
& 81.2 & \TFred{29.26}{-59.1} & \textbf{2.78}
& - & - & -
& - & - & - \\
& \texttt{ReASC} (online)
& 94.24 & \TFred{12.40}{-70.2} & \underline{7.60}
& 81.2 & \TFred{30.74}{-57.1} & \underline{2.64}
& 32.7 & \TFred{70.95}{-30.1} & \textbf{0.46}
& 36.55 & \TFred{39.67}{-48.0} & \textbf{0.92} \\
\midrule

\multirow{6}{*}{\textsc{Gemma-3-27B}}
& pass@1 & 95.60 & - & - & 77.6 & - & - & 37.5 & - & - & 35.53 & - & - \\
& SC (k=16)
& 97.04 & 166.93 & 0.58
& 82.6 & 291.36 & 0.28
& 42.9 & 585.84 & 0.07
& 45.69 & 368.97 & 0.12 \\
& ESC (w=4)
& 97.04 & \TFred{47.73}{-71.4} & 2.03
& 82.6 & \TFred{126.56}{-56.6} & 0.65
& 42.8 & \TFred{408.31}{-30.3} & 0.10
& 45.69 & \TFred{210.78}{-42.9} & 0.22 \\
& ASC
& 97.04 & \TFred{47.71}{-71.4} & 2.03
& 83.8 & \TFred{121.29}{-58.4} & 0.69
& 43.0 & \TFred{384.90}{-34.3} & \underline{0.11}
& 45.69 & \TFred{199.62}{-45.9} & \underline{0.23} \\
\cmidrule{2-14}
& \texttt{ReASC} (offline)
& 96.89 & \TFred{29.36}{-82.4} & \textbf{3.30}
& 83.6 & \TFred{101.32}{-65.2} & \textbf{0.83}
& - & - & -
& - & - & - \\
& \texttt{ReASC} (online)
& 97.04 & \TFred{35.32}{-78.8} & \underline{2.75}
& 83.6 & \TFred{100.62}{-65.5} & \textbf{0.83}
& 42.5 & \TFred{352.77}{-39.8} & \textbf{0.12}
& 47.21 & \TFred{161.68}{-56.2} & \textbf{0.29} \\

\bottomrule
\end{tabular}}
\caption{
Full comparison across mathematical and general reasoning benchmarks.
Metrics include accuracy (Acc), average TFLOPs (TF), and compute efficiency (Acc/TF).
The best Acc/TF is shown in bold, and the second best is underlined.
TF reduction percentages (shown in red) are reported relative to SC.
}
\label{tab:overall_full_results}
\vspace{-0.45cm}
\end{table*}
\subsubsection{Online Calibration}
\label{sec:online_calibration}
In the online setting, labeled validation data are unavailable.
Accordingly, confidence statistics are estimated from the confidence scores obtained during the single-sample generation in Stage~1 across all test instances.
Specifically, we compute the mean and standard deviation $(\mu,\sigma)$ of these confidence scores using the test instances as the calibration set, without requiring any additional inference.
\paragraph{Selecting the Online Gating Threshold.}
As in the offline setting, the gating threshold $\tau_{\mathrm{gate}}$ is designed to accept a single response only when its reliability is sufficiently high.
However, in the online setting, labeled validation data are unavailable, and the confidence distribution of correct responses cannot be directly observed.
To approximate the role of correctness information used in the offline setting, we model the distribution of unlabeled confidence scores using a Gaussian Mixture Model (GMM), treating correct and incorrect responses as latent Gaussian components.
This modeling choice is supported by model-selection criteria such as AIC and BIC \cite{akaike2003new, schwarz1978estimating}, which consistently favor a two-component mixture, indicating that the confidence distribution is well captured by the resulting bimodal fit (see Appendix~\ref{appendix:gmm}).

Given the fitted GMM, we estimate the gating threshold using two criteria.
First, we approximate the mean confidence of correct responses using the mean of the higher-confidence GMM component, which serves as a surrogate for $\mu_{\mathrm{correct}}$.
Second, to conservatively define an acceptance region in the absence of labels, we derive a posterior-based threshold $\tau_{\mathrm{post}}$ by identifying the smallest confidence value $t$ for which the mixture posterior exceeds a target accuracy score $p_{\mathrm{target}}$:

{\small
\begin{equation}
P(z = 1 \mid r = t)
=
\frac{
\pi_{1}\,\mathcal{N}(t;\mu_{1},\sigma_{1}^{2})
}{
\pi_{1}\,\mathcal{N}(t;\mu_{1},\sigma_{1}^{2})
+
\pi_{2}\,\mathcal{N}(t;\mu_{2},\sigma_{2}^{2})
}
\end{equation}
}
where $\pi_j$, $\mu_j$, and $\sigma_j^2$ denote the weight, mean, and variance of component $j$, and $z=1$ indicates membership in the higher-confidence component.
The final gating threshold is defined as the maximum of the two estimates.
The full online calibration procedure is summarized in Algorithm~\ref{alg:online-gating}.

\section{Experiments}
\subsection{Experimental Setup.}
\paragraph{Datasets and Baselines.}
We evaluate \texttt{ReASC} on four reasoning benchmarks spanning mathematical and general-domain reasoning: GSM8K \cite{cobbe2021training}, MATH500 \cite{lightman2023let}, Omni-Math \cite{gao2024omni}, and GPQA-Diamond \cite{rein2024gpqa}, which requires expert-level knowledge and multi-step reasoning.
We report reference results, including Pass@1 for single-sample performance and self-consistency (SC) \cite{wang2022self} with a fixed budget of $k$=16.
We further compare \texttt{ReASC} with several representative adaptive self-consistency baselines that dynamically adjust the sampling process.
ASC \cite{aggarwal2023let} uses a count-based Beta stopping rule.
ESC \cite{li2024escape} performs early stopping when all responses in a fixed context window of size 4 converge to the same answer.
For \texttt{ReASC}, we evaluate both offline and online settings, except for Omni-Math and GPQA-Diamond, where only online calibration is reported due to the absence of labeled validation sets.
Additional details on datasets and baselines can be found in Appendix~\ref{appendix:dataset_baseline}.
\paragraph{Implementation Details.}
We conduct experiments using five instruction-tuned language models from multiple families and scales, including LLaMA-3.2 (3B) \cite{grattafiori2024llama}, Qwen-2.5 (3B, 7B) \cite{yang2025qwen3}, and Gemma-3 (4B, 27B) \cite{team2025gemma}, covering model sizes from 3B to 27B.
For confidence calibration, we use a held-out set of $k$=128 instances, and both offline and online variants of \texttt{ReASC} adopt a target accuracy of $p_{\text{target}} = 0.9$, selected based on the accuracy-cost trade-off.
Adaptive stopping uses a confidence threshold of $C_{\text{threshold}} = 0.95$ for the confidence-weighted Beta update, following the default ASC setting, with a scaling factor of $\lambda = 0.7$.
We report accuracy, average inference cost (measured in TFLOPs), and their combined efficiency metric Acc/TF, which captures the trade-off between accuracy and computational cost, along with 95\% confidence intervals for accuracy (Appendix~\ref{appendix:confidence_interval}).
TFLOPs are estimated following \citet{kaplan2020scaling} by approximating the compute required to generate $2N$ FLOPs per token, where $N$ is the number of model parameters.
Detailed hyperparameter settings are provided in Appendix~\ref{appendix:more ablations}.

\begin{table}[t]
\centering
\renewcommand{\arraystretch}{1.05}
\resizebox{\linewidth}{!}{
\begin{tabular}{l
cc
cc}
\toprule
\multirow{2}{*}{\textbf{Model}}
& \multicolumn{2}{c}{\textbf{GSM8K}} &
  \multicolumn{2}{c}{\textbf{MATH500}} \\
\cmidrule(lr){2-3}
\cmidrule(lr){4-5}
& \textbf{Accept Ratio} & \textbf{Accept Acc}
& \textbf{Accept Ratio} & \textbf{Accept Acc} \\
\midrule

\multicolumn{5}{l}{\textbf{Offline}} \\
\textsc{LLaMA-3.2-3B}
& 48.98 & 91.33 
& 7.2 & 86.11 \\

\textsc{Gemma-3-4B}
& 51.18 & 97.78 
& 34.0 & 96.47 \\

\textsc{Qwen-2.5-7B}
& 59.59 & 97.58 
& 38.4 & 91.67 \\

\textsc{Gemma-3-27B}
& 60.58 & 98.62 
& 36.6 & 97.27 \\

\midrule

\multicolumn{5}{l}{\textbf{Online}} \\
\textsc{LLaMA-3.2-3B}
& 28.13 & 93.53 
& 17.6 & 88.28 \\

\textsc{Gemma-3-4B}
& 33.28 & 98.41 
& 28.2 & 98.58 \\

\textsc{Qwen-2.5-7B}
& 37.91 & 97.40 
& 31.8 & 93.08 \\

\textsc{Gemma-3-27B}
& 40.63 & 99.44 
& 36.2 & 97.31 \\

\bottomrule
\end{tabular}
}
\caption{
\textbf{Stage 1 acceptance ratio and accuracy.}
Stage 1 resolves a substantial fraction of inputs, while preserving accuracy under both offline and online calibration.
}
\label{tab:stage1_acceptance}
\vspace{-0.15cm}
\end{table}






\begin{figure}[t]
\centering
\includegraphics[width=\linewidth]{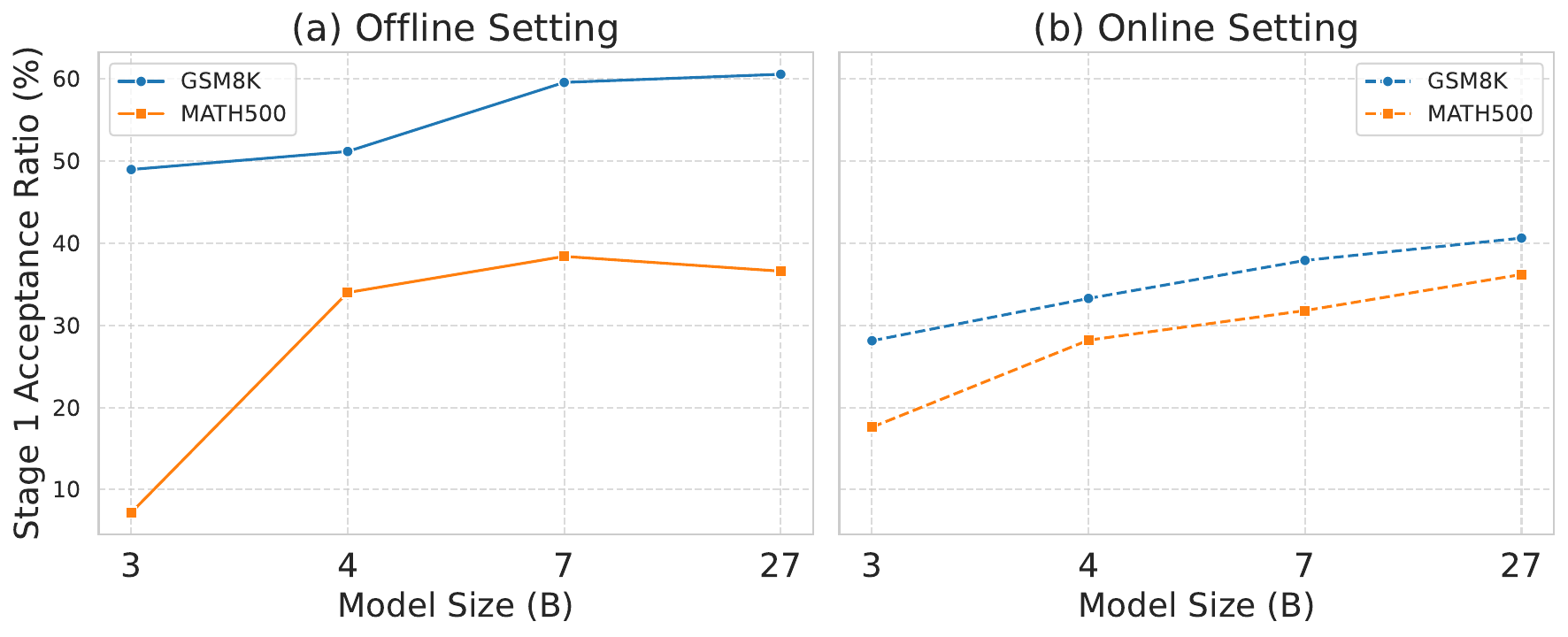}
\caption{
\textbf{Stage 1 acceptance ratio versus model size.}
Acceptance increases with model scale across datasets.}
\label{fig:stage1_ratio_vs_modelsize}
\vspace{-0.45cm}
\end{figure}
\subsection{Main Result.}
As shown in Table~\ref{tab:overall_full_results}, \texttt{ReASC} achieves the strongest accuracy-cost trade-off across all five models and four benchmarks, as measured by Acc/TF.
Specifically, \texttt{ReASC} consistently attains the lowest inference cost among self-consistency and adaptive baselines while preserving accuracy.
For example, on GSM8K with Gemma-3-4B, \texttt{ReASC} reduces inference cost by approximately 70\% relative to standard self-consistency, while consistently outperforming existing adaptive baselines in Acc/TF.
These improvements are observed consistently across both offline and online settings, highlighting that \texttt{ReASC} remains effective for practical deployment even without offline calibration.
Moreover, the same efficiency improvement persists across model scales ranging from 3B to 27B parameters, indicating the robustness and generalizability of \texttt{ReASC}.
Together, these results show the effectiveness of reliability-aware evidence accumulation by enabling efficient sampling decisions.

\subsection{Stage 1 Acceptance Analysis.}
We begin by examining whether \texttt{ReASC} can reliably identify instances for which evidence accumulation is unnecessary.
Specifically, we analyze the single-sample decision stage, which determines whether a reliable decision can be made from a single response.
We measure the Stage~1 acceptance ratio, defined as the fraction of instances resolved after a single response, and the accuracy of accepted instances.
As shown in Table~\ref{tab:stage1_acceptance}, the acceptance ratio consistently increases with model size across datasets, while acceptance accuracy mostly remains above 90\% under both offline and online calibration.
Figure~\ref{fig:stage1_ratio_vs_modelsize} further visualizes this trend, indicating that as model capability improves, Stage~1 reliably identifies instances that can be resolved from a single response.

\begin{table}[t]
\centering
\renewcommand{\arraystretch}{1.05}
\resizebox{\columnwidth}{!}{
\begin{tabular}{
l l
cc
cc
}
\toprule
& &
\multicolumn{2}{c}{\textbf{GSM8K}} &
\multicolumn{2}{c}{\textbf{MATH}} \\
\cmidrule(lr){3-4} \cmidrule(lr){5-6}
\textbf{Model} & \textbf{Method} &
\textbf{Acc $\uparrow$} & \textbf{TF $\downarrow$} &
\textbf{Acc $\uparrow$} & \textbf{TF $\downarrow$} \\
\midrule

\multirow{2}{*}{LLaMA-3.2-3B}
& ASC (count)  & 84.99 & 5.99 & 55.42 & 19.46 \\
& Ours (conf) & 84.25 & \textbf{3.95} & 54.80 & \textbf{14.75} \\

\midrule

\multirow{2}{*}{Qwen2.5-7B}
& ASC (count)  & 94.16 & 13.07 & 80.52 & 36.59 \\
& Ours (conf) & 94.04 & \textbf{10.00} & \textbf{81.49} & \textbf{28.22} \\

\midrule

\multirow{2}{*}{Gemma3-27B}
& ASC (count)  & 96.92 & 47.76 & 85.48 & 123.63 \\
& Ours (conf) & 96.53 & \textbf{28.02} & 85.17 & \textbf{101.25} \\

\bottomrule
\end{tabular}
}
\caption{
\textbf{Stage 2 performance after Stage 1 filtering.}
Reliability-aware accumulation reduces TF while preserving accuracy to count-based stopping.
}
\label{tab:after_stage1}
\vspace{-0.45cm}
\end{table}
\begin{figure*}[t]
\centering
\includegraphics[width=\linewidth]{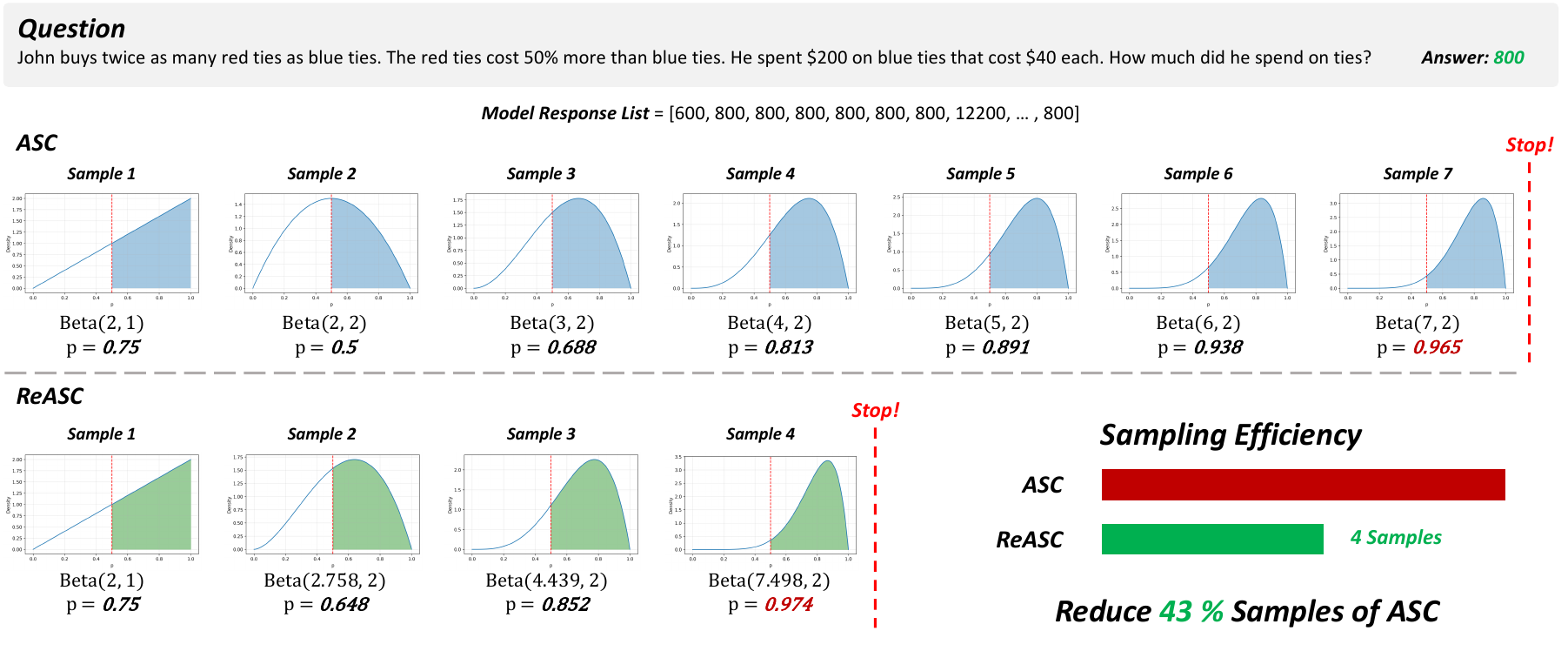}
\caption{
\textbf{Confidence-weighted update improves sampling efficiency.}
Each sampled response updates a Beta posterior, shown as the shaded region in each plot.
With sampling stopping at $p \geq 0.95$, ASC requires seven uniform updates, while \texttt{ReASC} reaches in four confidence-weighted updates, reducing sample cost by 43\%.
}
\label{fig:analysis}
\vspace{-0.45cm}
\end{figure*}
\subsection{Stage~2 Aggregation Analysis.}
We next examine whether reliability-aware evidence accumulation can efficiently identify sufficient evidence when a single response is insufficient.
To enable this analysis, we isolate the behavior of Stage~2 by reporting results only on instances not accepted at Stage~1 and comparing \texttt{ReASC} with ASC.
As shown in Table~\ref{tab:after_stage1}, reliability-aware accumulation still consistently reduces inference cost relative to count-based stopping while preserving comparable accuracy.
This suggests that while response counts provide a useful base signal, incorporating response-level confidence enables sufficient evidence to be identified with fewer samples when additional sampling is required.
\subsection{Stage-wise Ablation Studies.}
While the preceding analyses establish that each stage behaves as intended in isolation, they do not reveal how the two stages jointly contribute to the overall efficiency of \texttt{ReASC}.
We therefore conduct a stage-wise ablation to disentangle the complementary contributions of the two stages.
We compare three variants: \textbf{ASC}, which relies on count-based stopping; a \textbf{Stage~2 only} variant that applies confidence-weighted Beta update to all instances; and the full \textbf{\texttt{ReASC}} framework.
As shown in Table~\ref{tab:ablation_stage1}, replacing count-based stopping with reliability-aware accumulation reduces inference cost relative to ASC by identifying sufficient evidence earlier.
When comparing the Stage~2-only variant with \texttt{ReASC}, incorporating Stage~1 further reduces inference cost while preserving accuracy, indicating that evidence accumulation is unnecessary for a subset of instances in which a single response already provides reliable evidence.
Together, these results demonstrate that the two stages play complementary roles: Stage~2 improves the efficiency of evidence accumulation when sampling is required, while Stage~1 avoids unnecessary sampling by identifying instances where a single response already provides sufficient evidence.
\begin{table}[t]
\centering
\renewcommand{\arraystretch}{1.05}
\resizebox{\linewidth}{!}{
\begin{tabular}{l l
cc
cc}
\toprule
\textbf{Model} & \textbf{Method}
& \multicolumn{2}{c}{\textbf{GSM8K}}
& \multicolumn{2}{c}{\textbf{MATH500}} \\
\cmidrule(lr){3-4} \cmidrule(lr){5-6}
& & \textbf{Acc $\uparrow$} & \textbf{TF $\downarrow$}
& \textbf{Acc $\uparrow$} & \textbf{TF $\downarrow$} \\
\midrule

\multirow{3}{*}{\textsc{LLaMA-3.2-3B}}
& ASC
& 83.85 & 6.27
& 55.00 & 20.13 \\

& \texttt{ReASC} (Stage2 only)
& \textbf{84.38} & 5.33
& \textbf{55.20} & 18.76 \\

& \texttt{ReASC}
& 83.85 & \textbf{4.38}
& 55.00 & \textbf{18.27} \\

\midrule

\multirow{3}{*}{\textsc{Qwen2.5-7B}}
& ASC
& 94.24 & 13.40
& 80.80 & 37.25 \\

& \texttt{ReASC} (Stage2 only)
& \textbf{94.24} & 11.90
& \textbf{81.20} & 34.05 \\

& \texttt{ReASC}
& 94.09 & \textbf{10.43}
& 81.20 & \textbf{29.26} \\

\bottomrule
\end{tabular}
}
\caption{
\textbf{Stage-wise ablation of \texttt{ReASC}.}
Stage~2 yields comparable accuracy than count-based stopping, while Stage~1 primarily reduces inference cost when applied.}
\label{tab:ablation_stage1}
\vspace{-0.45cm}
\end{table}
\subsection{Confidence-Weighted Update Dynamics.}
We present a qualitative analysis of a representative GSM8K instance with LLaMA-3.2-3B-Instruct to demonstrate how confidence-weighted Beta updates affect evidence accumulation.
Figure~\ref{fig:analysis} visualizes the Beta posteriors of Adaptive Consistency (ASC) and \texttt{ReASC} under the same sampled responses.
ASC aggregates evidence uniformly, resulting in a gradual shift of the Beta distribution and stopping after seven samples.
In contrast, \texttt{ReASC} weights each update by response-level confidence, causing the posterior to concentrate more rapidly and reach the stopping threshold in four samples.
As a result, \texttt{ReASC} reduces sample cost by 43\% while converging to the same correct answer, enabling reliable decisions with fewer samples.

\section{Conclusion}
We present \texttt{ReASC}, a reliability-aware adaptive self-consistency framework that makes sampling decisions based on evidence sufficiency. 
By incorporating response-level reliability signals derived from model confidence, \texttt{ReASC} enables more effective evidence accumulation at test time.
Notably, \texttt{ReASC} demonstrates superior accuracy-cost trade-offs over existing adaptive sampling methods across models and datasets. 
Our findings highlight the importance of incorporating response reliability into adaptive sampling and suggest a principled direction for future work on efficient test-time sampling.

\section*{Limitations}
Our approach has several limitations. 
First, our current instantiation of \texttt{ReASC} estimates response-level reliability using model-derived confidence signals such as self-certainty.
This choice is supported by prior findings and further validated by our experiments across multiple model families and datasets; however, the calibration of confidence signals may still vary across models and tasks.
Second, as self-consistency is designed to elicit a model’s existing knowledge through multiple samples, \texttt{ReASC} leverages response-level confidence to make more efficient sampling decisions, treating higher confidence as more reliable reasoning.
While this assumption holds empirically across the benchmarks studied, it may be challenged in settings where models exhibit systematic overconfidence, suggesting that incorporating complementary reliability signals could further improve robustness.
Finally, our work focuses on inference-time adaptation without additional training, prioritizing simplicity and broad applicability.
While this design enables efficient deployment, incorporating learning-based approaches for reliability estimation could further improve accuracy and robustness, representing a promising direction for future work.


\section*{Acknowledgements}
This work was mainly supported by Institute of Information \& communications Technology Planning \& Evaluation (IITP) grant funded by the Korea government(MSIT) [No.RS-2023-00229780, Development of Artificial Intelligence Technology for Process-focused Evaluation(Student’s Learning Diagnosis; No.RS-2021-II211343, Artificial Intelligence Graduate School Program (Seoul National University)].
K. Jung is with ASRI, Seoul National University, Korea.

\bibliography{custom}
\newpage
\appendix
\section{Derivation of the Beta-Based Stopping Criterion}
\label{appendix:beta_derivation}

We provide a brief derivation of the probability expression used in the Beta-based stopping rule.  
Suppose that $V$ is the set of responses generated so far, and let $v_1$ and $v_2$ denote the number of samples assigned to the leading and second-best candidates.

Following the two-class reduction used in ASC, we consider the proportions $(p_1, p_2)$ associated with these two candidates under continued sampling.
Since $p_1 + p_2 = 1$, the event that the leading candidate remains dominant is equivalent to
\[
p_1 > p_2 \quad \Longleftrightarrow \quad p_1 > \tfrac{1}{2}.
\]

Under the Beta posterior induced by the accumulated counts in $V$, the distribution of $p_1$ is
\[
p_1 \sim \mathrm{Beta}(\alpha,\beta),
\qquad 
\alpha = v_1 + 1,\;\; \beta = v_2 + 1.
\]

The dominance probability of interest is therefore
\[
P(p_1 > p_2 \mid V) 
= 
P(p_1 > \tfrac{1}{2} \mid V).
\]

The Beta$(\alpha,\beta)$ density is
\[
f(t;\alpha,\beta)
=
\frac{1}{B(\alpha,\beta)}
t^{\alpha-1}(1-t)^{\beta-1},
\]
yielding
\[
P(p_1 > \tfrac{1}{2} \mid V)
=
\int_{1/2}^{1}
\frac{1}{B(\alpha,\beta)}
t^{\alpha-1}(1-t)^{\beta-1}\, dt.
\]

The regularized incomplete Beta function is defined as
\[
I_x(\alpha,\beta)
=
\frac{1}{B(\alpha,\beta)}
\int_{0}^{x}
t^{\alpha-1}(1-t)^{\beta-1}\, dt.
\]

Applying this definition with $x = 1/2$ and the identity
\[
\int_{1/2}^{1} f(t)\, dt
= 
1 - \int_0^{1/2} f(t)\, dt,
\]
we obtain
\[
P(p_1 > \tfrac{1}{2} \mid V)
=
1 - I_{1/2}(\alpha,\beta).
\]
which is the expression used in the stopping rule in Equation~\ref{eq:probability}

\section{AIC/BIC Analysis of Confidence Distributions}
\label{appendix:gmm}

To justify the use of a two-component Gaussian Mixture Model (GMM) in the online calibration setting, we evaluate how well GMMs with $n \in \{1,2,3,4\}$ components fit the unlabeled confidence-score distribution.
For each dataset, we fit GMMs via the EM algorithm and compute the Akaike Information Criterion (AIC) and Bayesian Information Criterion (BIC), which balance goodness of fit against model complexity (lower is better).
As shown in Table~\ref{table:aic-bic}, both AIC and BIC consistently select the two-component model across all settings, with clear improvements over a single-component model and no benefit from adding additional components.
These results indicate that the confidence distribution is well captured by a bimodal mixture, supporting our use of a two-component GMM for online threshold estimation.

\begin{table}[h]
\centering
\setlength{\tabcolsep}{6pt}
\renewcommand{\arraystretch}{1.1}
\resizebox{\columnwidth}{!}{
\begin{tabular}{llcccc}
\toprule
Model & Dataset & \multicolumn{4}{c}{\# Components $k$} \\
\cmidrule(lr){3-6}
 & & 1 & 2 & 3 & 4 \\
\midrule
\multirow{4}{*}{\textsc{Qwen2.5-3B}} 
  & \multirow{2}{*}{GSM8K} 
      & 5051.12 & \textbf{5004.18} & 5010.70 & 5016.93 \\
  &  & 5061.49 & \textbf{5030.10} & 5052.18 & 5073.96 \\
  \cmidrule{2-6}
  & \multirow{2}{*}{MATH500} 
      & 2385.12 & \textbf{2374.03} & 2376.92 & 2383.82 \\
  &  & 2398.55 & \textbf{2395.10} & 2410.64 & 2430.18 \\
\midrule
\multirow{4}{*}{\textsc{Gemma3-4B}} 
  & \multirow{2}{*}{GSM8K} 
      & 7381.64 & \textbf{7268.06} & 7276.34 & 7280.79 \\
  &  & 7392.01 & \textbf{7305.21} & 7309.53 & 7327.84 \\
  \cmidrule{2-6}
  & \multirow{2}{*}{MATH500} 
      & 2962.88 & \textbf{2950.26} & 2954.60 & 2950.96 \\
  &  & 2971.34 & \textbf{2971.31} & 2984.68 & 3005.46 \\
\bottomrule
\end{tabular}
}
\caption{AIC/BIC scores for GMM with $k$ components $(k=1,2,3,4)$. 
Bold indicates the best (lowest) value.}
\label{table:aic-bic}
\end{table}

\begin{algorithm}[t]
\small
\caption{Online Gating Threshold Calibration}
\label{alg:online-gating}
\begin{algorithmic}[1]

\REQUIRE Unlabeled test set $\{x_i\}_{i=1}^n$, target accuracy $p_{\mathrm{target}}$
\ENSURE Gating threshold $\tau_{\mathrm{gate}}$

\STATE \textcolor{gray}{// (1) Compute confidence for each instance}
\FOR{$i = 1$ to $n$}
    \STATE Generate response $y_i$
    \STATE Compute confidence $S(y_i)$
\ENDFOR

\STATE \textcolor{gray}{// (2) Fit GMM and compute surrogate correct mean}
\STATE Fit a 2-component GMM to $\{ S(y_i) \}$
\STATE Let component $c^\ast$ be the one with the larger mean
\STATE $\mu_{\mathrm{approx}} \gets \mathbb{E}_{r \sim c^\ast}[\, r \,]$ \; \textcolor{gray}{// surrogate correct mean}

\STATE \textcolor{gray}{// (3) Posterior-based threshold search}
\STATE Sort $\{S(y_i)\}$ in ascending order as candidate thresholds
\FOR{each threshold $t$}
    \STATE Compute $P(z=1 \mid r=t)$ under the fitted GMM
    \IF{$P(z=1 \mid r=t) \ge p_{\mathrm{target}}$}
        \STATE $\tau_{\mathrm{post}} \gets t$
        \STATE break
    \ENDIF
\ENDFOR

\STATE $\tau_{\mathrm{gate}} \gets \max(\mu_{\mathrm{approx}},\ \tau_{\mathrm{post}})$
\end{algorithmic}
\end{algorithm}

\section{Datasets and Baselines}
\label{appendix:dataset_baseline}

\subsection{Datasets}
We evaluate all methods on four reasoning benchmarks spanning mathematical and general-domain reasoning.
Table~\ref{table:dataset_description} illustrates the statistics and the corresponding license information for each dataset.
Below, we briefly describe each dataset and its evaluation protocol.

\paragraph{GSM8K.}
GSM8K is a grade-school-level mathematical reasoning benchmark consisting of 8,500 training and 1,319 test questions.
Each question requires multi-step arithmetic reasoning expressed in natural language.
Following standard practice, we evaluate accuracy based on the exact match of the final numerical answer extracted from the model output.

\paragraph{MATH500.}
MATH500 is a subset of the MATH benchmark designed to evaluate advanced mathematical problem-solving.
It covers a diverse range of topics, including algebra, geometry, number theory, and calculus.
We use the official test split of 500 problems and report the accuracy based on the verdict of the LLM Judge, namely xVerify \cite{chen2025xverify}.

\paragraph{Omni-Math.}
Omni-Math is a recently proposed large-scale benchmark for mathematical reasoning that emphasizes problem diversity and difficulty.
It includes questions that require longer reasoning chains and compositional mathematical skills.
We randomly sampled 1000 problems from the test set and report the accuracy based on the verdict of the LLM Judge, namely xVerify \cite{chen2025xverify}.

\paragraph{GPQA-Diamond.}
GPQA-Diamond is a general-domain reasoning benchmark curated to require expert-level knowledge and multi-step inference.
Questions span scientific and technical domains and are intentionally designed to be difficult for non-expert models.
We evaluate performance using exact-match accuracy against the answer choices.

\begin{table}[t]
\centering
\renewcommand{\arraystretch}{1.15}
\resizebox{\linewidth}{!}{
\begin{tabular}{lcccc}
\toprule
Dataset &Answer Format & $N_{q}$ &$L_{q}$ &License \\
\midrule
GSM8K &arabic number & 1319 & 239.9  & MIT License \\
MATH500 &arabic number & 500 & 195.9  & MIT License \\
Omni-Math &arabic number & 4428 & 270.8  & Apache-2.0 \\
GPQA-Diamond &option (A-D) & 197 & 598.1 & MIT License \\
\bottomrule
\end{tabular}
}
\caption{Relevant information of five datasets. $N_{q}$ denotes the number of questions in each dataset. $L_{q}$ denotes the average length of questions in each dataset.}
\label{table:dataset_description}
\end{table}

\subsection{LLM Inference Configuration.}
For all experiments, we perform inference using the default generation configurations recommended for each model, without additional tuning.
Specifically, LLaMA-3.2 is evaluated with temperature $0.6$ and top-$p$ $0.9$;
Qwen-2.5 uses temperature $0.7$, top-$p$ $0.8$, and top-$k$ $20$;
and Gemma-3 adopts temperature $1.0$, top-$p$ $0.95$, and top-$k$ $64$.
This design choice ensures that performance differences primarily reflect the behavior of adaptive sampling strategies rather than model-specific decoding heuristics.

\subsection{Baselines}
We compare \texttt{ReASC} against several representative inference-time baselines that differ in their sampling and stopping strategies.

\paragraph{Pass@1.}
Pass@1 reports the base model performance using a single sampled response.
This serves as a lower-bound reference for sampling-based methods.

\paragraph{Self-Consistency (SC).}
Self-Consistency aggregates multiple independently sampled reasoning trajectories and selects the most frequent final answer.
We use a fixed sampling budget of $k{=}16$ for all datasets.

\paragraph{Adaptive Consistency (ASC).}
ASC is an adaptive self-consistency method that dynamically determines when to stop sampling based on a count-based Beta stopping criterion.
All sampled responses are treated as equally informative, and sampling terminates once the Beta posterior exceeds $C_{threshold}=0.95$.

\paragraph{Early-Stopping Self-Consistency (ESC).}
ESC performs early stopping when the model produces identical answers within a fixed context window.
We use a window size $w$ of 4, as suggested in the original work.

\section{Analysis of Confidence Mapping Design}
\label{appendix:confidence_mapping}
We study how different confidence mapping functions affect the accuracy-cost trade-off in our adaptive sampling framework.
We consider the following alternative designs, each of which maps a normalized confidence score.

\paragraph{Mean-normalized aggregation.}
We consider a linear aggregation of normalized confidence scores,
\[
    v(y_i) \leftarrow v(y_i) + \frac{S(y_i)}{\mathbb{E}[\mathcal{D}_{val}]}.
\]
This mapping applies proportional vote increments without non-linear scaling.

\paragraph{Sigmoid-based mapping.}
Another alternative applies a sigmoid transformation,
\[
    v(y_i) \leftarrow v(y_i) + \sigma(\lambda z(y_i)),
\]
which compresses confidence scores into a bounded range.

\paragraph{Unbounded exponential mapping.}
We also evaluate an exponential mapping without a lower bound,
\[
    v(y_i) \leftarrow v(y_i) + \exp(\lambda z(y_i)).
\]

\paragraph{Proposed mapping.}
Finally, we propose an exponential mapping with a lower bound,
\[
    v(y_i) \leftarrow v(y_i) + \max(1, \exp(\lambda z(y_i))).
\]
\begin{table}[t]
\centering
\setlength{\tabcolsep}{3pt}
\renewcommand{\arraystretch}{1.1}
\resizebox{\columnwidth}{!}{
\begin{tabular}{l l cc}
\toprule
\textbf{Model} & \textbf{Mapping} & \textbf{GSM8K Acc/TF} & \textbf{MATH Acc/TF} \\
\midrule

\multirow{5}{*}{\textsc{LLaMA-3.2-3B}}
& ASC (count-based)            & 13.37 & 2.73 \\
\cmidrule{2-4}
& mean           & 13.92 & 2.86 \\
& \textit{sigmoid} & \textit{10.32} & \textit{2.60} \\
& exponential         & 14.74 & 2.84 \\
& \textbf{ours}  & \textbf{15.83} & \textbf{2.87} \\
\midrule

\multirow{5}{*}{\textsc{Qwen-2.5-7B}}
& ASC (count-based)            & 7.03 & 2.17 \\
\cmidrule{2-4}
& mean           & 6.93 & 2.14 \\
& \textit{sigmoid} & \textit{3.54} & \textit{1.55} \\
& exponential         & 6.60 & 2.12 \\
& \textbf{ours}  & \textbf{7.60} & \textbf{2.38} \\
\midrule

\multirow{5}{*}{\textsc{Gemma-3-27B}}
& ASC (count-based)            & 2.03 & 0.69 \\
\cmidrule{2-4}
& mean           & 2.04 & 0.67 \\
& \textit{sigmoid} & \textit{1.10} & \textit{0.42} \\
& exponential         & 2.04 & 0.67 \\
& \textbf{ours}  & \textbf{2.49} & \textbf{0.73} \\

\bottomrule
\end{tabular}
}
\caption{
\textbf{Analysis of Confidence Mapping Design.}
Results on LLaMA-3.2-3B, Qwen-2.5-7B, and Gemma-3-27B models.
The proposed mapping consistently achieves the best accuracy-compute trade-off.
}
\label{tab:confidence_mapping_design_representative}
\end{table}
\begin{figure*}[t]
\centering
\includegraphics[width=\linewidth]{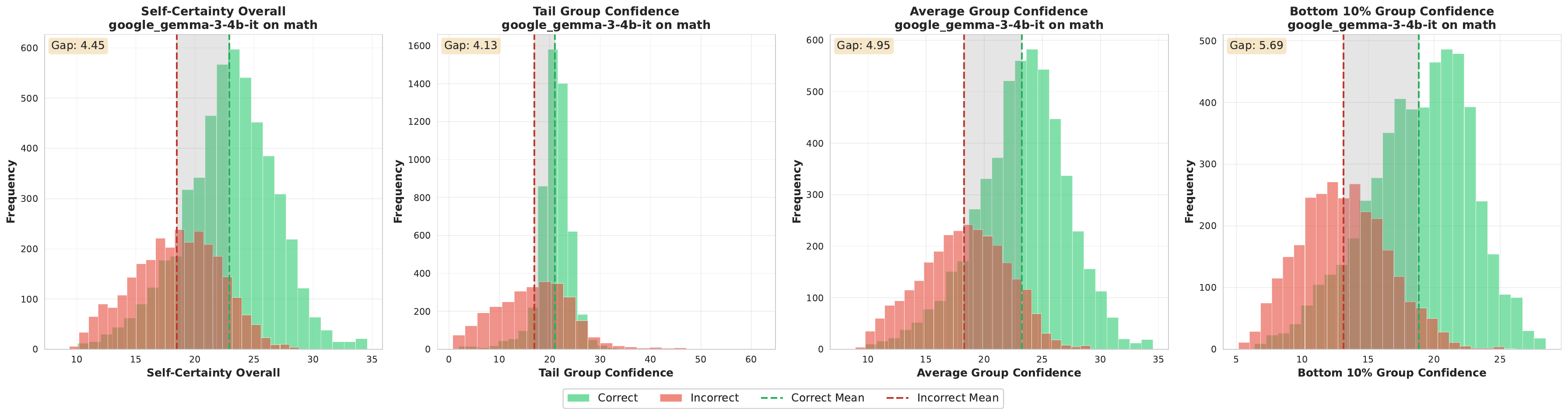}
\caption{
\textbf{Comparison of four self-certainty variants on a representative MATH instance using Gemma-3-4B-it.}
Among the variants, Bottom $10\%$ Group Confidence yields the largest gap between correct and incorrect responses, providing the clearest separation.}
\label{fig:confidence_design}
\end{figure*}
\noindent We also compared the alternative designs with ASC, which represents the count-based design.
To isolate the effect of confidence mapping, we report Acc/TF as the primary metric.
Results on three models (LLaMA-3.2-3B, Qwen-2.5-7B, and Gemma-3-27B) are shown in Table~\ref{tab:confidence_mapping_design_representative}.
While all variants achieve comparable accuracy, their efficiency differs substantially.
Sigmoid-based mappings compress confidence scores to the range [0, 1], leading to slower vote accumulation and higher computational cost.
In contrast, exponential mappings better differentiate high-confidence responses, enabling earlier stopping.
Among them, introducing a lower bound yields the most stable improvement across model scales, leading to the strongest accuracy-cost trade-off.

\begin{table}[t]
\centering
\small
\renewcommand{\arraystretch}{1.15}
\resizebox{\linewidth}{!}{
\begin{tabular}{lcc}
\toprule
\textbf{Confidence Metric} & \textbf{AUROC} & \textbf{Gap} \\
\midrule
Response-level Self-Certainty        & 0.801 & 4.45 \\
Tail Group Confidence           & 0.699 & 4.13 \\
Average Group Confidence        & 0.823 & 4.95 \\
\textbf{Bottom $10\%$ Group Confidence}  & \textbf{0.860} & \textbf{5.69} \\
\bottomrule
\end{tabular}
}
\caption{AUROC and gap between correct and incorrect means comparison of confidence metrics for distinguishing correct and incorrect responses.}
\label{tab:confidence_auroc}
\vspace{-0.45cm}
\end{table}
\section{Analysis on Confidence Score Design}
\label{appendix:confidence_design}
\subsection{Impact of Confidence Score}
We analyze several confidence metrics to assess how reliably they distinguish correct from incorrect responses, including response-level self-certainty, tail-group confidence, average-group confidence, and Bottom $10\%$ Group Confidence based on the confidence design choice from \citet{fu2025deep}.
We evaluate each metric using two complementary criteria: (i) the separation gap between the mean confidence of correct and incorrect responses, and (ii) the area under the ROC curve (AUROC), which measures ranking-based discriminative performance.
As shown in Figure~\ref{fig:confidence_design}, Bottom $10\%$ Group Confidence exhibits the most significant separation gap, indicating clearer distributional separation between correct and incorrect responses.
Consistent with this observation, AUROC results reported in Table~\ref{tab:confidence_auroc} show that Bottom $10\%$ Group Confidence achieves the strongest discriminative performance among the compared confidence metrics.
Together, these results support our choice of Bottom $10\%$ Group Confidence as the confidence signal in \texttt{ReASC}.

\subsection{Sensitivity to Group Size.}
We further analyze the sensitivity of the Bottom $10\%$ Group Confidence to the sliding-window group size by evaluating its discriminative performance using AUROC.
As shown in Table~\ref{tab:group_size_auroc}, AUROC remains relatively stable over a wide range of group sizes across both Gemma-3-4B-it and Qwen-2.5-3B-it on the MATH dataset, indicating that the metric is not overly sensitive to this hyperparameter.
Among the tested values, a group size of 128 consistently yields the strongest or near-strongest separation between correct and incorrect responses.
Based on this robustness-performance trade-off, we use a group size of 128 in all experiments.
\begin{table}[t]
\centering
\small
\renewcommand{\arraystretch}{1.15}
\resizebox{\linewidth}{!}{
\begin{tabular}{lcc}
\toprule
& \multicolumn{2}{c}{\textbf{AUROC}} \\
\cmidrule{2-3}
\textbf{Window Size} & \textbf{Gemma-3-4B-it} & \textbf{Qwen-2.5-3B-it} \\
\midrule
32   & 0.832 & 0.714 \\
64   & 0.853 & 0.728 \\
\textbf{128}  & \textbf{0.874} & \textbf{0.744} \\
256  & 0.874 & 0.741 \\
512  & 0.867 & 0.736 \\
768  & 0.862 & 0.725 \\
\bottomrule
\end{tabular}
}
\caption{AUROC sensitivity of Bottom $10\%$ Group Confidence to sliding window group size on the MATH.}
\label{tab:group_size_auroc}
\vspace{-0.45cm}
\end{table}

\begin{table*}[t]
\centering
\footnotesize
\renewcommand{\arraystretch}{1.05}
\setlength{\tabcolsep}{6pt}
\resizebox{\linewidth}{!}{
\begin{tabular}{llcccc}
\toprule
\multirow{2}{*}{\textbf{Model}} &
\multirow{2}{*}{\textbf{Method}} &
\multicolumn{1}{c}{\textbf{GSM8K}} &
\multicolumn{1}{c}{\textbf{MATH500}} &
\multicolumn{1}{c}{\textbf{Omni-Math}} &
\multicolumn{1}{c}{\textbf{GPQA-Diamond}} \\
& &
\multicolumn{1}{c}{\textbf{95\% CI}} &
\multicolumn{1}{c}{\textbf{95\% CI}} &
\multicolumn{1}{c}{\textbf{95\% CI}} &
\multicolumn{1}{c}{\textbf{95\% CI}} \\
\midrule

\multirow{5}{*}{\textsc{LLaMA-3.2-3B-Instruct}} &
Maj@k (k=16) & [81.96, 85.90] & [50.40, 59.20] & [13.81, 18.42] & [19.29, 31.47] \\
& ESC (w=4) & [81.96, 85.90] & [50.40, 59.20] & [13.81, 18.42] & [18.78, 30.46] \\
& ASC & [81.88, 85.82] & [50.60, 59.60] & [13.61, 18.12] & [19.29, 31.47] \\
\cmidrule{2-6}
& \texttt{ReASC} (offline) & [81.80, 85.82] & [50.60, 59.20] & -- & -- \\
& \texttt{ReASC} (online) & [81.73, 85.75] & [49.20, 58.20] & [13.01, 17.42] & [19.80, 31.98] \\
\midrule

\multirow{5}{*}{\textsc{Qwen-2.5-3B-Instruct}} &
Maj@k (k=16) & [87.87, 91.05] & [68.60, 76.20] & [24.42, 30.03] & [24.86, 37.56] \\
& ESC (w=4) & [87.72, 90.98] & [68.60, 76.20] & [24.42, 30.03] & [22.84, 36.04] \\
& ASC & [87.72, 90.98] & [68.60, 76.20] & [24.62, 30.13] & [24.86, 37.56] \\
\cmidrule{2-6}
& \texttt{ReASC} (offline) & [87.88, 90.75] & [68.00, 75.60] & -- & -- \\
& \texttt{ReASC} (online) & [87.72, 90.98] & [67.80, 75.40] & [24.72, 30.23] & [23.86, 37.06] \\
\midrule

\multirow{5}{*}{\textsc{Gemma-3-4B-it}} &
Maj@k (k=16) & [90.67, 93.48] & [67.60, 75.60] & [27.13, 32.83] & [23.86, 36.56] \\
& ESC (w=4) & [90.60, 93.48] & [67.60, 75.60] & [27.13, 32.93] & [23.86, 36.56] \\
& ASC & [90.67, 93.48] & [67.60, 75.40] & [27.43, 33.14] & [23.86, 36.56] \\
\cmidrule{2-6}
& \texttt{ReASC} (offline) & [90.52, 93.40] & [67.40, 75.40] & -- & -- \\
& \texttt{ReASC} (online) & [90.60, 93.48] & [67.20, 75.00] & [27.63, 33.33] & [23.35, 36.56] \\
\midrule

\multirow{5}{*}{\textsc{Qwen-2.5-7B-Instruct}} &
Maj@k (k=16) & [93.03, 95.53] & [77.20, 84.00] & [30.23, 36.24] & [29.44, 43.15] \\
& ESC (w=4) & [92.95, 95.38] & [77.20, 84.00] & [30.23, 36.24] & [28.43, 41.62] \\
& ASC & [92.95, 95.38] & [77.40, 84.20] & [30.43, 36.44] & [29.44, 43.15] \\
\cmidrule{2-6}
& \texttt{ReASC} (offline) & [92.80, 95.30] & [77.40, 84.20] & -- & -- \\
& \texttt{ReASC} (online) & [92.80, 95.30] & [78.00, 84.60] & [30.03, 36.24] & [29.44, 43.15] \\
\midrule

\multirow{5}{*}{\textsc{Gemma-3-27B-it}} &
Maj@k (k=16) & [96.06, 97.95] & [79.40, 85.80] & [40.04, 46.05] & [38.58, 52.79] \\
& ESC (w=4) & [96.06, 97.95] & [79.40, 85.80] & [39.94, 45.85] & [31.98, 45.69] \\
& ASC & [96.06, 97.95] & [80.60, 86.60] & [40.04, 46.05] & [38.58, 52.79] \\
\cmidrule{2-6}
& \texttt{ReASC} (offline) & [95.91, 97.80] & [80.20, 86.40] & -- & -- \\
& \texttt{ReASC} (online) & [96.06, 97.95] & [80.20, 86.40] & [40.04, 45.85] & [40.09, 54.31] \\
\bottomrule
\end{tabular}
}
\caption{95\% confidence intervals for accuracy computed via non-parametric bootstrap over test instances.}
\label{tab:accuracy_ci}
\end{table*}
\section{Confidence Interval Analysis}
\label{appendix:confidence_interval}
To assess the statistical robustness of the reported accuracy, we compute 95\% confidence intervals using a non-parametric bootstrap over test instances.
For each method, we collect the verdict for each test instance and generate 2,000 bootstrap resamples by sampling instances with replacement.
Accuracy is computed for each resample, and the 95\% confidence interval is obtained from the 2.5 and 97.5 percentiles of the resulting distribution.
\begin{table}[t]
\centering
\small
\renewcommand{\arraystretch}{1.15}
\resizebox{\columnwidth}{!}{
\begin{tabular}{l c c c c  c c c}
\toprule
& & \multicolumn{3}{c}{\textbf{Qwen2.5-3B}} & \multicolumn{3}{c}{\textbf{Gemma-3-4B}} \\
\cmidrule(lr){3-5} \cmidrule(lr){6-8}
\textbf{Regime} & $\boldsymbol{p_{\text{target}}}$ 
& \textbf{Acc} & \textbf{TF} & \textbf{Acc/TF}
& \textbf{Acc} & \textbf{TF} & \textbf{Acc/TF} \\
\midrule
\multirow{3}{*}{Offline}
& 0.9  & 72.0 & 17.19 & \textbf{4.19} & 71.8 & 27.18 & \textbf{2.64} \\
& 0.95 & 72.4 & 17.37 & 4.17 & 72.0 & 28.41 & 2.53 \\
& 0.99 & 72.8 & 20.01 & 3.64 & 72.0 & 29.41 & 2.45 \\
\midrule
\multirow{3}{*}{Online}
& 0.9  & 72.4 & 18.06 & \textbf{4.01} & 71.8 & 27.70 & \textbf{2.59} \\
& 0.95 & 72.0 & 18.79 & 3.83 & 71.8 & 27.94 & 2.57 \\
& 0.99 & 72.8 & 19.97 & 3.65 & 72.0 & 28.74 & 2.51 \\
\bottomrule
\end{tabular}
}
\caption{\textbf{Sensitivity analysis of the target accuracy $p_{\text{target}}$ for \texttt{ReASC}.} Results are reported in terms of accuracy (Acc), average TFLOPs (TF), and their efficiency ratio (Acc/TF) under both offline and online regimes.}
\label{tab:ptarget_sensitivity}
\end{table}

\section{More Ablation Studies.}
\label{appendix:more ablations}
\subsection{Selecting target accuracy}
\label{appendix:selecting_ptarget}
We conduct a hyperparameter sensitivity analysis on the target confidence threshold $p_{\text{target}}$ using the Math500 dataset.
Experiments are performed with two representative models, Qwen2.5-3B-Instruct and Gemma-3-4B-it, under both offline and online regimes of \texttt{ReASC}.
We vary $p_{\text{target}}$ across \{0.9, 0.95, 0.99\} and evaluate the resulting trade-off between accuracy and computational cost.
As shown in Table~\ref{tab:ptarget_sensitivity}, higher values of $p_{\text{target}}$ generally lead to increased sampling and higher inference cost, while providing only marginal accuracy improvements.
Across both models and regimes, $p_{\text{target}}=\textbf{0.9}$ consistently achieves the best accuracy–compute trade-off, and we therefore adopt this setting in all main experiments.

\subsection{Analysis of Calibration set size}
\begin{figure}[t]
\centering
\includegraphics[width=\linewidth]{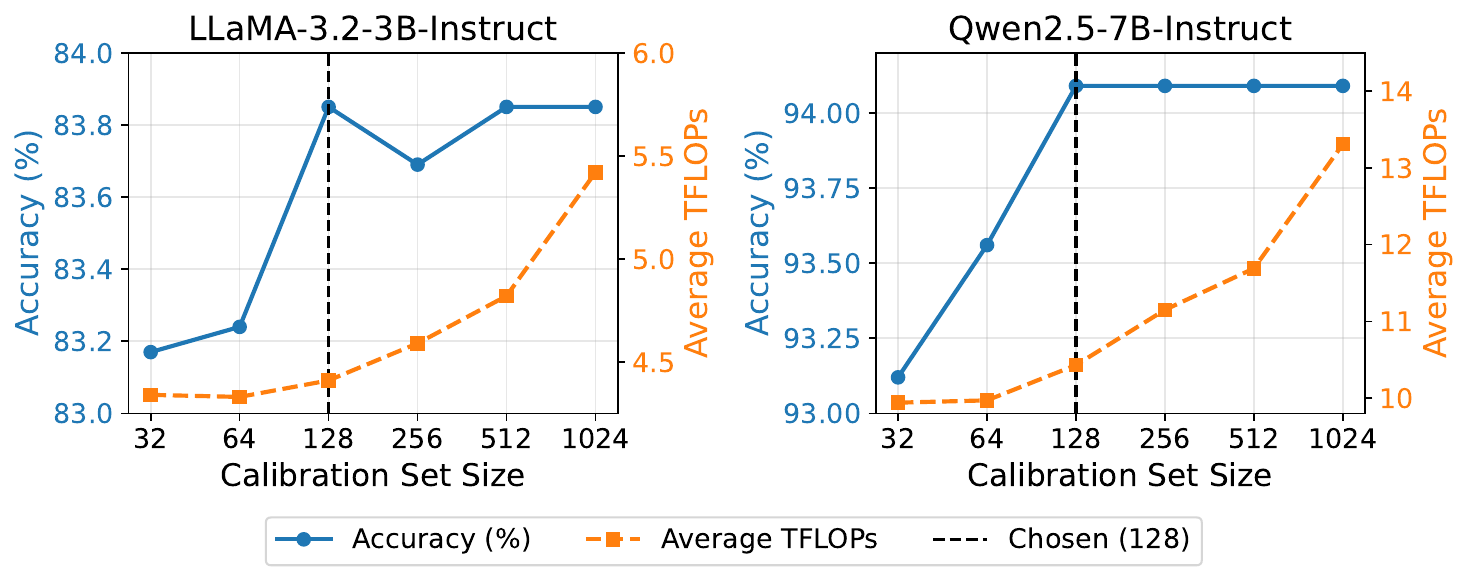}
\caption{
\textbf{Accuracy (left) and average TFLOPs (right) as a function of calibration set size.}
Accuracy gains diminish beyond a calibration size of 128, whereas inference cost increases steadily with larger calibration sets.
Based on this trade-off, we use a calibration size of 128 in all experiments.
}
\label{fig:calib_size}
\vspace{-0.45cm}
\end{figure}

We study the sensitivity of \texttt{ReASC} to the size of the calibration set used for threshold selection.
Figure~\ref{fig:calib_size} shows accuracy and average TFLOPs as a function of calibration set size for LLaMA-3.2-3B-Instruct and Qwen-2.5-7B-Instruct.
Across both models, accuracy improves with calibration size up to 128 examples and then saturates, showing negligible differences for larger sets.
In contrast, the average TFLOPs increase monotonically as the calibration size grows, reflecting higher calibration overhead.
These results indicate that a calibration size of 128 achieves the best trade-off between accuracy and inference cost, and we therefore use this value throughout our experiments.

\begin{table*}[t]
\centering
\footnotesize
\renewcommand{\arraystretch}{1.15}
\resizebox{\linewidth}{!}{
\begin{tabular}{lccccccccc}
\toprule
& \multicolumn{3}{c}{\textbf{StrategyQA}} & \multicolumn{3}{c}{\textbf{Letter}} & \multicolumn{3}{c}{\textbf{NQ-Open}} \\
\cmidrule(lr){2-4} \cmidrule(lr){5-7} \cmidrule(lr){8-10}
\textbf{Method} &
\textbf{Acc $\uparrow$} & \textbf{TF $\downarrow$} & \textbf{Acc/TF $\uparrow$} &
\textbf{Acc $\uparrow$} & \textbf{TF $\downarrow$} & \textbf{Acc/TF $\uparrow$} &
\textbf{Acc $\uparrow$} & \textbf{TF $\downarrow$} & \textbf{Acc/TF $\uparrow$} \\
\midrule
Maj@k ($k$=16)   & 63.87 & 29.05 & 2.20 & 44.6 & 10.02 & 4.45 & 26.8 & 21.28 & 1.26 \\
ESC ($w$=4)      & 63.87 &  9.22 & 6.93 & 44.0 &  5.69 & 7.73 & 26.8 & 17.93 & 1.49 \\
ASC              & 63.87 &  9.14 & 6.99 & 44.2 &  5.48 & 8.07 & 27.4 & 17.34 & 1.58 \\
\midrule
\texttt{ReASC} (online) & 64.09 & 9.08 & \textbf{7.06} & 45.6 & 4.70 & \textbf{9.70} & 27.0 & 15.04 & \textbf{1.80} \\
\bottomrule
\end{tabular}
}
\caption{\textbf{Results beyond math-focused reasoning benchmarks using Qwen2.5-7B-Instruct.} \texttt{ReASC} maintains the strongest accuracy-compute trade-off (Acc/TF) across StrategyQA, Last Letter Concatenation, and NQ-Open, suggesting that the benefit of reliability-aware adaptive sampling is not limited to math-focused reasoning.}
\label{tab:general_domain_results}
\end{table*}

\subsection{Selecting $\lambda$ for Confidence-Weighted Updates.}
\begin{figure}[h]
\centering
\includegraphics[width=\linewidth]{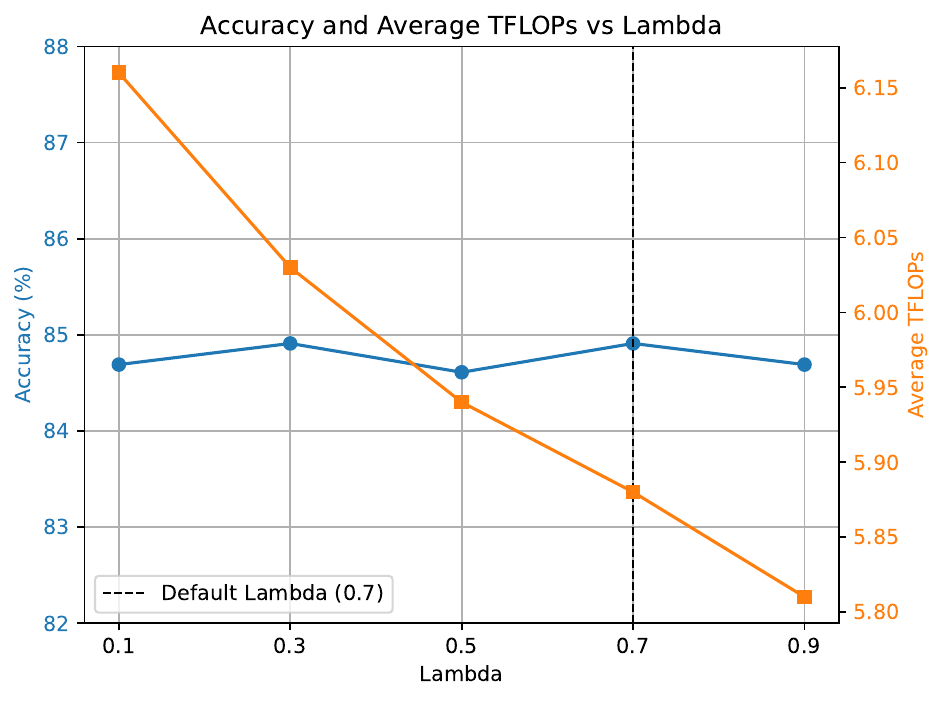}
\caption{
\textbf{Accuracy (left) and average TFLOPs (right) versus $\lambda$ on GSM8K with LLaMA-3.2-3B-Instruct.}
While TFLOPs decrease with larger $\lambda$, accuracy remains stable and peaks at $\lambda=0.7$, which we set as the default.
}
\label{fig:lambda_config}
\vspace{-0.45cm}
\end{figure}

We conduct a sensitivity study on the confidence-weighting parameter $\lambda$ using the GSM8K dataset with the LLaMA-3.2-3B-Instruct model.
We evaluate $\lambda \in \{0.1, 0.3, 0.5, 0.7, 0.9\}$ and report both accuracy and average TFLOPs.
In Figure~\ref{fig:lambda_config}, the average TFLOPs consistently decrease as $\lambda$ increases, indicating more aggressive evidence accumulation.
Accuracy peaks at $\lambda=0.7$, while larger values yield marginal cost reductions at the expense of accuracy.
Considering both accuracy and computation, we select $\lambda=0.7$ as it provides the most favorable accuracy–cost trade-off.

\subsection{Evaluation Beyond Math-Focused Reasoning}
While our main experiments focus on mathematical reasoning benchmarks, we additionally evaluate \texttt{ReASC} on more general-domain tasks to assess whether the benefit of reliability-aware adaptive sampling extends beyond math-focused settings.
Specifically, we consider StrategyQA\cite{geva2021did} for commonsense reasoning, Last Letter Concatenation\cite{wei2022chain} for symbolic manipulation, and NQ-Open\cite{lee2019latent} for open-domain question answering, using Qwen2.5-7B-Instruct and the same evaluation protocol as in our main experiments.
As shown in Table~\ref{tab:general_domain_results}, \texttt{ReASC} consistently achieves the strongest accuracy-compute trade-off across all three tasks.
In particular, \texttt{ReASC} maintains comparable or improved accuracy while requiring fewer TFLOPs than adaptive baselines, yielding the best Acc/TF in every setting.
These results suggest that the advantage of reliability-aware evidence accumulation extends beyond mathematical reasoning to broader reasoning and open-ended generation tasks.

\subsection{Confidence Reliability and Overconfident Errors}
We further examine whether the confidence signal used in \texttt{ReASC} provides a meaningful estimate of response reliability.
Using Qwen2.5-7B-Instruct, we partition sampled responses into five quantile-based confidence bins and measure empirical accuracy within each bin. 
Table~\ref{tab:confidence_bin_accuracy} shows that accuracy increases monotonically with confidence.
This indicates that higher self-certainty is generally associated with a higher likelihood of correctness.
Although overconfident errors remain possible, the overall trend supports the use of self-certainty as a practical reliability signal in \texttt{ReASC}.
\begin{table}[t]
\centering
\small
\setlength{\tabcolsep}{5pt}
\renewcommand{\arraystretch}{1.15}
\resizebox{\columnwidth}{!}{
\begin{tabular}{lcc}
\toprule
\textbf{Bin (quantile)} & \textbf{Confidence range} & \textbf{Accuracy within bin} \\
\midrule
1 (bottom 20\%) & [4.74, 7.52]   & 20.00\% \\
2 (20--40\%)    & [7.52, 10.30]  & 40.21\% \\
3 (40--60\%)    & [10.30, 13.08] & 51.92\% \\
4 (60--80\%)    & [13.08, 15.86] & 82.94\% \\
5 (top 20\%)    & [15.86, 18.64] & 93.27\% \\
\bottomrule
\end{tabular}}
\caption{Empirical accuracy across confidence bins on Qwen2.5-7B-Instruct. Accuracy increases monotonically with confidence, while high-confidence errors remain relatively infrequent.}
\label{tab:confidence_bin_accuracy}
\end{table}

\subsection{Practical Efficiency Under Batched Generation}
We further examine whether the practical efficiency of \texttt{ReASC} is limited by its stop-and-check mechanism.
While the default implementation includes a sequential component, \texttt{ReASC} can also be combined with batched generation by sampling multiple responses in parallel and then applying the same confidence-aware stopping rule.
To verify this, we evaluate a batched variant of \texttt{ReASC} on GSM8K using Qwen2.5-3B-Instruct.
As shown in Table~\ref{tab:batched_latency_results}, the batched variant substantially reduces latency compared to the sequential version, while preserving the same accuracy.
Although batch generation slightly increases TFLOPs, it still yields a favorable latency-compute trade-off relative to adaptive baselines. These results suggest that the practical efficiency of \texttt{ReASC} is not limited to purely sequential settings and extends to moderate batch-parallel serving regimes.
\begin{table}[t]
\centering
\renewcommand{\arraystretch}{1.15}
\resizebox{\columnwidth}{!}{
\begin{tabular}{lcccc}
\toprule
\textbf{Method} & \textbf{Batch size} & \textbf{Accuracy} & \textbf{TFLOPs} & \textbf{Latency} (s) \\
\midrule
ASC              & 1 & 89.39 & 7.57 & 17419.70 \\
ESC              & 4 & 89.39 & 8.03 &  8325.90 \\
\texttt{ReASC}   & 1 & 89.54 & 6.43 & 14124.74 \\
\texttt{ReASC} (batched) & 4 & 89.54 & 7.75 & \textbf{5509.05} \\
\bottomrule
\end{tabular}
}
\caption{Latency comparison under sequential and batched generation with Qwen2.5-3B-Instruct.}
\label{tab:batched_latency_results}
\vspace{-0.45cm}
\end{table}

\section{Use of AI Tools}
During the preparation of this paper, AI tools (e.g., OpenAI's ChatGPT) were used in a limited, supporting capacity. Specifically, they assisted in enhancing the clarity and fluency of the text and in suggesting relevant keywords during the writing process. All conceptual ideas, experimental designs, implementations, analyses, and final interpretations were developed entirely by the authors. The authors independently verified all cited references, and no citation was included solely based on AI-generated content. No private, unpublished, or sensitive information was shared with AI tools beyond what is explicitly described in this paper.

\end{document}